\documentclass[runningheads]{llncs}
\usepackage{float}
\usepackage{graphicx}
\usepackage{booktabs}   %
\usepackage{subcaption} %
\usepackage{amsmath}
\usepackage{amssymb}
\usepackage{amsfonts}
\newcommand{\citet}[1]{\cite{#1}}
\newcommand{\citep}[1]{\cite{#1}}
\usepackage{thmtools, thm-restate}

\usepackage{mathtools}
\DeclarePairedDelimiter\abs{\lvert}{\rvert}%

\usepackage{pgf,tikz,pgfplots}
\usepackage{mathrsfs}
\usetikzlibrary{arrows, decorations.markings}
\tikzstyle{vecArrow} = [thick, decoration={markings,mark=at position
   1 with {\arrow[semithick]{open triangle 60}}},
   double distance=1.4pt, shorten >= 5.5pt,
   preaction = {decorate},
   postaction = {draw,line width=1.4pt, white,shorten >= 4.5pt}]
\pgfplotsset{compat=1.13}

\usepackage{xcolor}
\definecolor{COCcolor}{HTML}{1F77B4}
\definecolor{WLcolor}{HTML}{FDB863}
\definecolor{SLcolor}{HTML}{E66101}
\definecolor{WRcolor}{HTML}{B2ABD2}
\definecolor{SRcolor}{HTML}{5E3C99}

\usepackage{prettyref}
\newcommand{\pref}{\prettyref}
\newrefformat{thm}{Theorem~\ref{#1}}
\newrefformat{cor}{Corollary~\ref{#1}}
\newrefformat{lem}{Lemma~\ref{#1}}
\newrefformat{cha}{Chapter~\ref{#1}}
\newrefformat{sec}{Section~\ref{#1}}
\newrefformat{app}{Appendix~\ref{#1}}
\newrefformat{tab}{Table~\ref{#1}}
\newrefformat{fig}{Figure~\ref{#1}}
\newrefformat{alg}{Algorithm~\ref{#1}}
\newrefformat{exa}{Example~\ref{#1}}
\newrefformat{def}{Definition~\ref{#1}}
\newrefformat{li}{Line~\ref{#1}}
\newrefformat{eq}{Equation~\ref{#1}}

\usepackage[ruled,vlined,linesnumbered]{algorithm2e}
\newcommand{\eqdef}{\buildrel \mbox{\tiny\textrm{def}} \over =}

\newcommand{\restrsname}{restriction domains of interest}

\newcommand{\restr}[2]{{#1}_{\restriction#2}}
\newcommand{\meta}[1]{{\color{blue}\textbf{[}#1\textbf{]}}}

\newcommand{\extendname}{\textsc{Extend}}

\newcommand{\relu}{\textsc{ReLU}}

\newcommand{\npost}[1]{\textsc{Post}}

\newcommand{\trimmedacas}[1]{\includegraphics[width=\linewidth]{#1}}
\DeclareMathOperator*{\argmax}{arg\,max}

\DeclareMathOperator*{\DPPre}{DeepPoly}

\SetKw{returnKw}{return}
\SetKw{breakKw}{break}
\SetKw{continueKw}{continue}
\SetKwFunction{Queue}{ConstructQueue}
\SetKwFunction{Push}{Push}
\SetKwFunction{Pop}{Pop}
\SetKwFunction{Sign}{Sign}
\SetKwFunction{Next}{Next}
\SetKwFunction{Append}{Append}
\SetKwFunction{Dim}{Dim}
\SetKwFunction{CWVert}{CWVert}
\SetKwFunction{Vert}{Vert}
\SetKwFunction{Hull}{ConvexHull}
\SetKwFunction{OrthantSign}{OrthantSign}
\SetKwFunction{PlaneIntersect}{HyperplaneIntersections}
\SetKwFunction{OrthantIntersect}{EnumerateOrthantIntersections}
\SetKwFunction{SplitPlane}{SplitPlane}
\SetKwFunction{AffineFunctionOn}{AffineFunctionOn}

\newcommand{\subsubsubsection}[1]{\smallskip\noindent\textbf{\emph{#1}}\enspace}

\renewcommand{\hat}[1]{\widehat{#1}}
\newcommand{\hatr}[2]{\widehat{\restr{#1}{#2}}}

\newcommand{\syrenn}{SyReNN}

\usepackage{ifthen}
\newcommand\target{arxiv}
\newcommand\onlyfor[3]{\ifthenelse{\equal{#1}{\target}}{#2}{#3}}

\usepackage{marvosym}

\makeatletter
\RequirePackage[bookmarks,unicode,colorlinks=true]{hyperref}%
   \def\@citecolor{blue}%
   \def\@urlcolor{blue}%
   \def\@linkcolor{blue}%

\def\orcidID#1{\smash{\href{http://orcid.org/#1}{\protect\raisebox{-1.25pt}{\protect\includegraphics{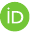}}}}}
\makeatother

\onlyfor{tacas}{
\usepackage[firstpage]{draftwatermark}
\SetWatermarkText{\hspace*{4.5in}\raisebox{6.1in}{\includegraphics[scale=0.1]{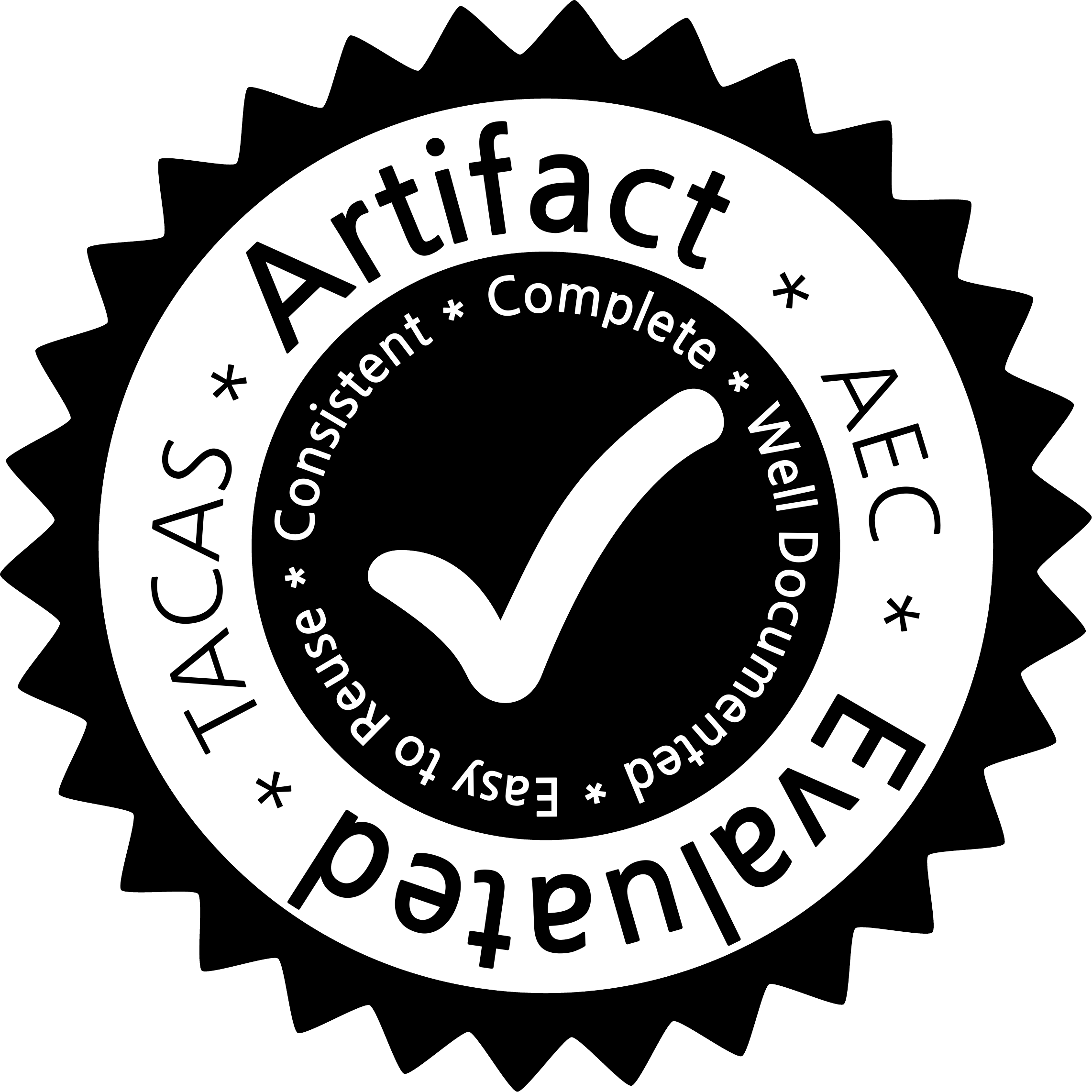}}}
\SetWatermarkAngle{0}
}{}

\usetikzlibrary{patterns}
\tikzset{
dot/.style = {circle, fill, minimum size=#1,
              inner sep=0pt, outer sep=0pt},
}

\begin{document}
\title{\syrenn: A Tool for Analyzing Deep Neural
Networks
\thanks{Artifact available at \url{https://zenodo.org/record/4124489}.}}
\author{Matthew Sotoudeh \orcidID{0000-0003-2060-1009} (\Letter)
\and Aditya V.\ Thakur \orcidID{0000-0003-3166-1517} (\Letter)}
\authorrunning{Sotoudeh and Thakur}
\institute{University of California, Davis CA 95616, USA\\
\email{\{masotoudeh,avthakur\}@ucdavis.edu}}
\maketitle              %
\begin{abstract}
    Deep Neural Networks (DNNs) are rapidly gaining popularity in a variety of
    important domains. Formally, DNNs are complicated vector-valued functions
    which come in a variety of sizes and applications. Unfortunately, modern
    DNNs have been shown to be vulnerable to a variety of attacks and buggy
    behavior. This has motivated recent work in formally analyzing the
    properties of such DNNs. This paper introduces \syrenn{}, a tool for
    understanding and analyzing a DNN by computing its \emph{symbolic
    representation.} The key insight is to decompose the DNN into
    linear functions. Our tool is designed for analyses using
    \emph{low-dimensional subsets} of the input space, a unique design point in
    the space of DNN analysis tools.
    We describe the tool and the underlying theory, then evaluate its use and
    performance on three case studies: computing Integrated Gradients,
    visualizing a DNN's decision boundaries, and patching a DNN.

\keywords{Deep Neural Networks  \and Symbolic representation \and Integrated Gradients}
\end{abstract}

\section{Introduction}
\label{sec:Introduction} 

Deep Neural Networks (DNNs)~\cite{Goodfellow:DeepLearning2016} have become the
state-of-the-art in a variety of applications including image
recognition~\cite{Szegedy:CVPR2016,Krizhevsky:CACM2017} and natural language
processing~\cite{BERT:CoRR2018}. Moreover, they are increasingly used in safety-
and security-critical applications such as autonomous
vehicles~\cite{julian2018deep} and medical diagnosis
\cite{Ching:RSIF2017,DBLP:journals/bib/MiottoWWJD18,Hosny:NatureReviews2018,Mendelson:AJR2019}.
These advances have been accelerated by improved hardware and algorithms.

DNNs (\pref{sec:Preliminaries}) are programs that compute a vector-valued
function, i.e., from $\mathbb{R}^n$ to $\mathbb{R}^m$. They are straight-line
programs written as a concatenation of alternating linear and non-linear
\emph{layers}. The coefficients of the linear layers are learned from data via
\emph{gradient descent} during a training process. A number of different
non-linear layers (called \emph{activation functions}) are commonly used,
including the \emph{rectified linear} and \emph{maximum pooling} functions.

Owing to the variety of application domains as well as deployment constraints,
DNNs come in many different sizes. For instance, large image-recognition and
natural-language processing models are trained and deployed using cloud
resources~\citep{Krizhevsky:CACM2017,BERT:CoRR2018}, medium-size models could
be trained in the cloud but deployed on hardware with limited
resources~\cite{julian2018deep}, and finally small models could be trained and
deployed directly on edge devices
\cite{sharma2016high,chen2019deep,DBLP:conf/pldi/GopinathGSS19,shiftry,kusupati2018fastgrnn}.
There has also been a recent push to compress trained models to reduce their
size~\cite{compression}. Such smaller models play an especially important
role in privacy-critical applications, such as wake word detection for voice
assistants, because they allow sensitive user data to stay on the user's own
device instead of needing to be sent to a remote computer for processing.

Although DNNs are very popular, they are not perfect. One particularly
concerning development is that modern DNNs have been shown to be extremely
vulnerable to \emph{adversarial examples,} inputs which are intentionally
manipulated to appear unmodified to humans but become misclassified by the
DNN~\citep{Szegedy:ICLR2014,Goodfellow:ICLR2015,DeepFool:CVPR2016,carlini2018audio}.
Similarly, \emph{fooling examples} are inputs that look like random noise to
humans, but are classified with high confidence by
DNNs~\citep{Nguyen:CVPR2015}. Mistakes made by DNNs have led to loss of life
\cite{teslacrash,ubercrash} and wrongful
arrests~\cite{translationarrest,wronglyaccused}. For this reason, it is
important to develop techniques for analyzing, understanding, and repairing~DNNs.

This paper introduces \syrenn{}, a tool for understanding and analyzing DNNs.
\syrenn{} implements state-of-the-art algorithms for computing precise
symbolic representations of piecewise-linear DNNs (\pref{sec:Background}). Given
an input subspace of a DNN, \syrenn{} computes a symbolic representation that
decomposes the behavior of the DNN into finitely-many linear functions.
\syrenn{} implements the one-dimensional analysis algorithm of Sotoudeh and
Thakur~\cite{DBLP:conf/nips/SotoudehT19} and extends it to the two-dimensional
setting as described in~\pref{sec:Computing}.

\subsubsubsection{Key insights.}
There are two key insights enabling this approach, first identified in
Sotoudeh and Thakur~\cite{DBLP:conf/nips/SotoudehT19}. First, most popular DNN
architectures today are \emph{piecewise-linear}, meaning they can be precisely
decomposed into finitely-many linear functions. This allows us to reduce their
analysis to equivalent questions in linear algebra, one of the most
well-understood fields of modern mathematics. Second, many applications only require
analyzing the behavior of the DNN on a \emph{low-dimensional subset} of the
input space. Hence, whereas prior work has attempted to give up precision for
efficiency in analyzing high-dimensional input regions
\cite{singh2018fast,Singh:POPL2019,ai2:SP2018}, our work has focused on
algorithms that are \emph{both efficient and precise} in analyzing
lower-dimensional regions (\pref{sec:Computing}).

\subsubsubsection{Tool design.} 
The \syrenn{} tool  is designed to be easy to use and extend, as well as
efficient (\pref{sec:Tool}). The core of \syrenn{} is written as a
highly-optimized, parallel C++ server using Intel TBB for parallelization
\cite{reinders2007intel} and Eigen for matrix operations \cite{eigenweb}. A
user-friendly Python front-end interfaces with the PyTorch
deep learning framework \cite{PyTorch}. 

\subsubsubsection{Use cases.}
We demonstrate the utility of \syrenn{} using three applications. The first
computes \emph{Integrated Gradients} (IG), a state-of-the-art measure used to
determine which input dimensions (e.g., pixels for an image-recognition
network) were most important in the final classification produced by the
network (\pref{sec:IntegratedGradients}). The second precisely visualizes the
decision boundaries of a DNN (\pref{sec:Visualization}). The last
\emph{patches} (repairs) a DNN to satisfy some desired specification involving
infinitely-many points (\pref{sec:Patching}). Thus, we believe that \syrenn{}
is an interesting and useful tool in the toolbox for understanding and
analyzing DNNs.

\subsubsubsection{Contributions.} The contributions of this paper are:
\begin{itemize} 
    \item A definition of symbolic representation of DNNs (\pref{sec:Background}).
    \item An efficient algorithm for computing symbolic representations for
        DNNs over low-dimensional input subspaces (\pref{sec:Computing}).
    \item A design of a usable and well-engineered tool implementing these
        ideas called \syrenn{} (\pref{sec:Tool}).
    \item Three applications of \syrenn{} (\pref{sec:Applications}).
\end{itemize}

\pref{sec:Preliminaries} presents preliminaries about DNNs;
\pref{sec:Related} presents related work; \pref{sec:Conclusion} concludes.
SyReNN is available on GitHub at \url{https://github.com/95616ARG/SyReNN}.

\section{Preliminaries}
\label{sec:Preliminaries}

We now formally define the notion of \emph{DNN} we will use in this paper.
\begin{definition}
    A \emph{Deep Neural Network} (DNN) is a function $f : \mathbb{R}^n \to
    \mathbb{R}^m$ which can be written $f = f_1 \circ f_2 \cdots \circ f_n$ for
    a sequence of \emph{layer functions} $f_1$, $f_2$, \ldots, $f_n$.
\end{definition}

Our work is primarily concerned with the popular class of
\emph{piecewise-linear} DNNs, defined below.  In this definition and the rest
of this paper, we will use the term ``polytope'' to mean a \emph{convex and
bounded} polytope except where specified.

\begin{definition}
    \label{def:PWL}
    A function $f : \mathbb{R}^n \to \mathbb{R}^m$ is \emph{piecewise-linear (PWL)}
    if its input domain $\mathbb{R}^n$ can be partitioned into finitely-many
    possibly-unbounded polytopes $X_1, X_2, \ldots, X_k$ such that
    $\restr{f}{X_i}$ is \emph{linear} for every $X_i$.
\end{definition}

The most common activation function used today is the ReLU function, a PWL
activation function which is defined below.

\begin{definition}
    \label{def:ReLU}
    The \emph{rectified linear function} (ReLU) is a function $\relu :
    \mathbb{R}^n \to \mathbb{R}^m$ defined component-wise by
    \[
        \relu(\vec{v})_i \coloneqq
        \begin{cases}
            0 &\text{if } v_i < 0 \\
            v_i &\text{otherwise}, \\
        \end{cases}
    \]
    where $\relu(\vec{v})_i$ is the $i$th component of the vector
    $\relu(\vec{v})$ and $v_i$ is the $i$th component of the vector $\vec{v}$.
\end{definition}

In order to see that $\relu{}$ is PWL, we must show that its input domain
$\mathbb{R}^n$ can be partitioned such that, in each partition, $\relu{}$ is
linear. In this case, we can use the orthants of $\mathbb{R}^n$ as our
partitioning: within each orthant, the signs of the components do not change
hence $\relu{}$ is the linear function that just zeros out the negative
components.

Although we focus on $\relu{}$ due to its popularity and expository power,
\syrenn{} works with a number of other popular PWL layers include MaxPool,
Leaky ReLU, Hard Tanh, Fully-Connected, and Convolutional layers, as defined
in~\cite{Goodfellow:DeepLearning2016}. PWL layers have become exceedingly
common. In fact, nearly all of the state-of-the-art image recognition models
bundled with Pytorch~\cite{paszke2017automatic} are PWL.

\begin{example}
    The DNN $f : \mathbb{R}^1 \to \mathbb{R}^1$ defined by
    \[
        f(x) \coloneqq
        \begin{bmatrix}
            1 & -1 & -1
        \end{bmatrix}
        \relu{}\left(
            \begin{bmatrix}
                1 & -1 \\
                1 & 0 \\
                -1 & 0
            \end{bmatrix}
            \begin{bmatrix}
                x \\ 1
            \end{bmatrix}
        \right)
    \]
    can be broken into layers $f = f_1 \circ f_2 \circ f_3$ where
    \[
        f_1(x)
        \coloneqq
        \begin{bmatrix}
            1 & -1 \\
            1 & 0 \\
            -1 & 0
        \end{bmatrix}
        \begin{bmatrix}
            x \\ 1
        \end{bmatrix},
        \quad
        f_2 = \relu{},
        \quad
        \text{and}
        \quad
        f_3(\vec{v}) =
        \begin{bmatrix}
            1 & -1 & -1
        \end{bmatrix}\vec{v}.
    \]
    The DNN's input-output behavior on the domain $[-1, 2]$ is shown
    in~\pref{fig:example_dnn}.
\end{example}

\section{A Symbolic Representation of DNNs} 
\label{sec:Background}

\begin{figure}[t]
    \centering
    \begin{tikzpicture}[scale=0.5]
        \begin{axis}[ymin=-1.3,ymax=0.3,xlabel={Input $x$},ylabel={Output $y$},font=\huge,
            max space between ticks=100]
            \addplot[ultra thick,red,domain=-1:0,samples=2] {x};
            \addplot[ultra thick,blue,domain=0:1,samples=2] {-x};
            \addplot[ultra thick,green,domain=1:2,samples=2] {-x + (x - 1)};
        \end{axis}
    \end{tikzpicture}
    \caption{ 
        Example function for which $\hatr{f}{[-1, 2]} = \{ [-1, 0], [0, 1], [1,
        2] \}$.}
    \label{fig:example_dnn}
\end{figure}
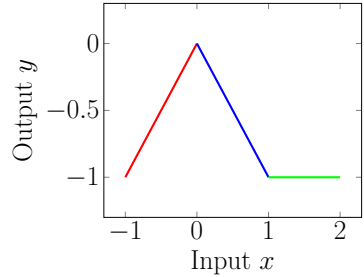

We formalize the symbolic representation according to the following definition:
\begin{definition}
    \label{def:fhat}
    Given a PWL function $f : \mathbb{R}^n \to \mathbb{R}^m$ and a bounded
    convex polytope $X \subseteq \mathbb{R}^n$, we define the \emph{symbolic
    representation of $f$ on $X$}, written $\hatr{f}{X}$, to be a finite set of
    polytopes $\hatr{f}{X}~=~\{ P_1, \ldots, P_n \}$, such that:
    \begin{enumerate}
        \item The set $\{ P_1, P_2, \ldots, P_n \}$ partitions $X$, except
            possibly for overlapping boundaries.
        \item Each $P_i$ is a bounded convex polytope.
        \item Within each $P_i$, the function $\restr{f}{P_i}$ is linear.
    \end{enumerate}
\end{definition}
Notably, if $f$ is a DNN using only PWL layers, then $f$ is PWL and so we can
define $\hatr{f}{X}$.
This symbolic representation allows one to reduce questions about the DNN $f$ to
questions about finitely-many linear functions $F_i$. For example, because
linear functions are convex, to verify that $\forall x \in X.\quad f(x) \in Y$
for some polytope $Y$, it suffices to verify $\forall P_i \in \hatr{f}{X}.
\forall \vec{v} \in \Vert(P_i).  \quad f(\vec{v}) \in Y$, where $\Vert(P_i)$ is
the (finite) set of vertices for the bounded convex polytope $P_i$; thus, here
both of the quantifiers are over finite sets. The symbolic representation
described above can be seen as a generalization of the \textsc{ExactLine}
representation~\citep{DBLP:conf/nips/SotoudehT19}, which considered only
\emph{one-dimensional} restriction domains of interest.

\begin{example}
    Consider again the DNN $f : \mathbb{R}^1 \to \mathbb{R}^1$ given by
    \[
        f(x) \coloneqq
        \begin{bmatrix}
            1 & -1 & -1
        \end{bmatrix}
        \mathrm{ReLU}\left(
            \begin{bmatrix}
                1 & -1 \\
                1 & 0 \\
                -1 & 0
            \end{bmatrix}
            \begin{bmatrix}
                x \\ 1
            \end{bmatrix}
        \right)
    \]
    and the region of interest $X = [-1, 2]$. The input-output behavior of $f$
    on $X$ is shown in~\pref{fig:example_dnn}. From this, we can see that
    \[
        \hatr{f}{X} = \{ [-1, 0], [0, 1], [1, 2] \}.
    \]
    Within each of these partitions, the input-output behavior is
    \emph{linear,} which for $\mathbb{R}^1 \to \mathbb{R}^1$ we can see
    visually as just a line segment. As this set fully partitions $X$, then,
    this is a valid $\hatr{f}{X}$.
\end{example}

\section{Computing the Symbolic Representation}
\label{sec:Computing}

This section presents an efficient algorithm for computing $\hatr{f}{X}$ for a
DNN $f$ composed of PWL layers. To retain both scalability and precision, we
will \emph{require the input region $X$ be two-dimensional}. This design choice
is relatively unexplored in the neural-network analysis literature (most
analyses strike a balance between precision and scalability, ignoring
dimensionality). We show that, for two-dimensional~$X$, we can use an efficient
polytope representation to produce an algorithm that demonstrates good
best-case and in-practice efficiency while retaining full precision. This
algorithm represents a direct generalization of the approach of
\citet{DBLP:conf/nips/SotoudehT19}.

The difficulties our algorithm addresses arise from three areas. First, when
computing $\hatr{f}{X}$ there may be exponentially many such partitions on all
of $\mathbb{R}^n$ but only a small number of them may intersect with $X$.
Consequently, the algorithm needs to be able to find those partitions that
intersect with $X$ efficiently without explicitly listing all of the partitions
on $\mathbb{R}^n$. Second, it is often more convenient to specify the
partitioning via \emph{hyperplanes separating the partitions} than explicit
polytopes. For example, for the one-dimensional $\relu{}$ function we may
simply state that the line $x = 0$ separates the two partitions, because
$\relu{}$ is linear both in the region $x \leq 0$ and $x \geq 0$. Finally,
neural networks are typically composed of sequences of linear and
piecewise-linear layers, where the partitioning imposed by each layer
individually may be well-understood but their composition is more complex. For
example, identifying the linear partitions of $y = \relu{}(4\cdot\relu{}(-3x - 1) +
2)$ is non-trivial, even though we know the linear partitions of each composed
function individually.

Our algorithm only requires the user to specify the hyperplanes defining the
partitioning for the activation function used in each layer; our current
implementation comes with support for common PWL activation functions. For
example, if a $\relu{}$ layer is used for an $n$-dimensional input vector, then
the hyperplanes would be defined by the equations $x_1 = 0, x_2 = 0, \ldots,
x_n = 0$. It then computes the symbolic representation for a \emph{single layer
at a time}, composing them sequentially to compute the symbolic representation
across the entire network.

To allow such compositions of layers, instead of directly
computing $\hatr{f}{X}$, we will define another primitive, denoted by the
operator $\otimes$ and sometimes referred to as
$\extendname$, such that
\begin{equation}
    \extendname{}(h, \hat{g}) = h \otimes \hat{g} = \hat{h \circ g}.
\end{equation}
Consider $f = f_n \circ f_{n-1} \circ
\cdots \circ f_1$, and let $I : x \mapsto x$ be the identity map. 
$I$ is linear across its entire input space, and, thus, $ \hatr{I}{X} = \{ X \}$.
By the definition of $\extendname{}(f_1, \cdot)$, we have
$
    f_1 \otimes \hatr{I}{X} = \hatr{(f_1 \circ I)}{X} = \hatr{f_1}{X}
$,
where the final equality holds by the definition of the identity map $I$.
We can then iteratively apply this procedure to inductively compute
$\hatr{(f_i \circ \cdots \circ f_1)}{X}$ from $\hatr{(f_{i-1} \circ \cdots
f_1)}{X}$ like so:
\[
    f_i \otimes \hatr{(f_{i - 1} \circ \cdots \circ f_1)}{X} = \hatr{(f_i \circ
    f_{i - 1} \circ \cdots \circ f_1)}{X}
\]
until we have computed $\hatr{(f_n \circ f_{n-1} \circ \cdots \circ f_1)}{X} = \hatr{f}{X}$, 
which is the required symbolic representation.

\subsection{Algorithm for \extendname}
\label{sec:Algorithm} 

\pref{alg:relu-2d-extend} present an algorithm for computing $\extendname$ for
arbitrary PWL functions, where $\extendname{}(h, \hat{g}) = h \otimes \hat{g} =
\hat{h \circ g}$. 

\subsubsubsection{Geometric intuition for the algorithm.}
\label{sec:AlgCaseStudy}
Consider the $\relu{}$ function (\pref{def:ReLU}).
It can be shown that, within any orthant (i.e., when the signs of all
coefficients are held constant), $\relu{}(\vec{x})$ is equivalent to some
linear function, in particular the element-wise product of $\vec{x}$ with a
vector that zeroes out the negative-signed components. However, for our
algorithm, all we need to know is that the linear partitions of $\relu{}$
(in this case the orthants) are separated by hyperplanes $x_1 = 0, x_2 = 0,
\ldots, x_n = 0$.

Given a two-dimensional convex bounded polytope $X$, the execution of the
algorithm for $f = \relu{}$ can be visualized as follows. We pick some vertex
$v$ of $X$, and begin traversing the boundary of the polytope in
counter-clockwise order. If we hit an orthant boundary (corresponding to some
hyperplane $x_i = 0$), it implies that the behavior of the function behaves
differently at the points of the polytope to one side of the boundary from
those at the other side of the boundary. Thus, we \emph{partition $X$ into
$X_1$ and $X_2$}, where $X_1$ lies to one side of the hyperplane and $X_2$ lies
to the other side. We recursively apply this procedure to $X_1$ and $X_2$ until
the resulting polytopes all lie on exactly one side of every hyperplane
(orthant boundary). But lying on exactly one side of every hyperplane (orthant
boundary) implies each polytope lies entirely within a linear partition of the
function (a single orthant), hence the application of the function on that
polytope is linear, and hence we have our partitioning.

\subsubsubsection{Functions used in algorithm.}
\label{sec:AlgDefinitions}
Given a two-dimensional bounded convex polytope $X$, \Vert{$X$} returns a list
of its vertices in counter-clockwise order, repeating the initial vertex at the
end.
Given a set of points $X$, \Hull{$X$} represents their convex hull (the
smallest bounded polytope containing every point in $X$).
Given a scalar value $x$, \Sign{$x$} computes the sign of that value (i.e.,
$-1$ if $x < 0$, $+1$ if $x > 0$, and $0$ if $x = 0$).

\subsubsubsection{Algorithm description.}
\label{sec:Algorithms-ReLU2D}
The key insight of the algorithm is to recursively
partition the polytopes until such a partition lies entirely within a linear
region of the function $f$. \pref{alg:relu-2d-extend} begins by constructing a
queue containing the polytopes of $\hatr{g}{X}$. Each iteration either removes
a polytope from the queue that lies entirely in one linear region (placing it in $Y$), or splits
(partitions) some polytope into two smaller polytopes that get put back into
the queue. When we pop a polytope $P$ from the queue, \pref{li:IterPlanes}
iterates over all hyperplanes $N_k\cdot x = b_k$ defining the piecewise-linear
partitioning of $f$, looking for any for which some vertex $V_i$ lies on the
positive side of the hyperplane and another vertex $V_j$ lies on the negative
side of the hyperplane. If none exist (\pref{li:NoPlanes}), by convexity we are
guaranteed that the entire polytope lies entirely on one side with respect to
every hyperplane, meaning it lies entirely within a linear partition of $f$.
Thus, we can add it to $Y$ and continue. If two such vertices are found
(starting~\pref{li:SomePlane}), then we can find ``extreme'' $i$ and $j$
indices such that $V_i$ is the last vertex in a counter-clockwise traversal to
lie on the same side of the hyperplane as $V_1$ and $V_j$ is the last vertex
lying on the opposite side of the hyperplane. We then call $\SplitPlane{}$ (\pref{alg:split-plane}) to
actually partition the polytope on opposite sides of the hyperplane, adding
both to our worklist.

\begin{figure}[t]
{\small
\begin{algorithm}[H]
    \DontPrintSemicolon
    \KwIn{$\hatr{g}{X} = \{ P_1, \ldots, P_n \}$.}
    \KwOut{$\hatr{f \circ g}{X}$}
    $W \gets$ \Queue{$\hatr{g}{X}$}\;
    $Y \gets \emptyset$\;\label{li:DefineY}
    \While{$W$ not empty}{
        $P \gets$ \Pop{$W$}\;
        $V \gets \Vert{$P$}$\;
        $K \gets \{ N_k \mid \exists i, j: \Sign(N_k\cdot g(V_i) - b_k) > 0 \wedge \Sign(N_k\cdot g(V_j) - b_k) < 0 \}$\;\label{li:IterPlanes}
        \If{$K = \emptyset$}{\label{li:NoPlanes}
            $Y \gets Y \cup \{ P \}$\;
            \continueKw{}
        }
        $N, b \gets$ any element from $K$\;\label{li:SomePlane}
        $i \gets \argmax_i \{ \Sign(N \cdot g(V_i) - b) = \Sign(N \cdot g(V_1) - b) \}$\;
        $j \gets \argmax_j \{ \Sign(N \cdot g(V_j) - b) \neq \Sign(N \cdot g(V_i) - b) \}$\;
        \For{$V' \in \SplitPlane(V, g, i, j, N, b)$}{
            $W \gets$ \Push{$W, \Hull(V')$}\;
        }
    }
    \returnKw{$Y$}
    \caption{$f \otimes \hatr{g}{X}$ for two-dimensional $X$. $f$ is defined by
    hyperplanes $N_1 \cdot x = b_1$ through $N_m \cdot x = b_m$ such that,
    within any partition imposed by the hyperplanes $f$ is equivalent to some
    affine function.}
    \label{alg:relu-2d-extend}
\end{algorithm}
}
\end{figure}
\begin{figure}[t]
{\small
\begin{algorithm}[H]
    \DontPrintSemicolon
    \KwIn{$V$, the vertices of the polytope in the input space of $g$.
    A function $g$.
    $i$ is the index of the last vertex lying on the same side of the orthant
    face as $V_1$.
    $j$ is the index of the last vertex lying on the opposite side of the
    orthant face as $V_1$.
    $N$ and $b$ define the hyperplane $N \cdot x = b$ to split on.}
    \KwOut{$\{ P_1, P_2 \}$, two sets of vertices whose convex hulls form a
    partitioning of $V$ such that each lies on only one side of the $N \cdot x
    = b$ hyperplane.}
    $p_i \gets V_i + \frac{b - N\cdot g(V_i)}{N\cdot (g(V_{i+1}) - g(V_{i}))}(V_{i+1} - V_i)$\;
    $p_j \gets V_j + \frac{b - N\cdot g(V_j)}{N\cdot (g(V_{j+1}) - g(V_{j}))}(V_{j+1} - V_j)$\;
    $A \gets \{ p_i, p_j \} \cup \{ v \in V \mid \Sign(N\cdot v - b) = \Sign(N\cdot V_i - b) \}$\;
    $B \gets \{ p_i, p_j \} \cup \{ v \in V \mid \Sign(N\cdot v - b) = \Sign(N\cdot V_j - b) \}$\;
    \returnKw{$\{ A, B \}$}
    \caption{SplitPlane$(V, g, i, j, N, b)$}
    \label{alg:split-plane}
\end{algorithm}
}
\end{figure}

In the best case, each partition is in a single orthant:
the algorithm never calls \SplitPlane{} at all --- it merely iterates over all
of the $n$ input partitions, checks their $v$ vertices, and appends to the
resulting set (for a best-case complexity of $O(nv)$). In the worst case, it
splits each polytope in the queue on each face, resulting in exponential time
complexity. As we will show in~\pref{sec:Applications}, this
exponential worst-case behavior is not encountered in practice, thus making
\syrenn{} a practical tool for DNN analysis.

\onlyfor{arxiv}{
    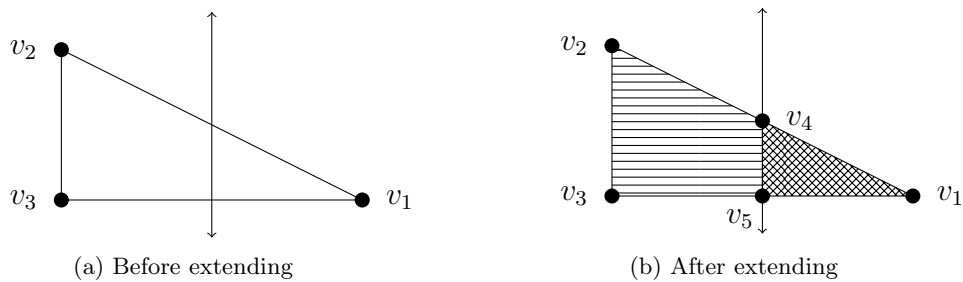
\begin{figure}[H]
    \centering
    \begin{subfigure}[t]{0.4\linewidth}
        \begin{tikzpicture}
            \large
            \draw[draw=none] (2, 0) -- (-2, 0) -- (-2, 2) -- cycle;
            \draw (2, 0) -- (-2, 0);
            \draw (-2, 0) -- (-2, 2);
            \draw (-2, 2) -- (2, 0);

            \draw[<->] (0, -0.5) -- (0, 2.5);

            \draw node[dot=2mm] at (2, 0) {};
            \draw node at (2.5, 0) {$v_1$};

            \draw node[dot=2mm] at (-2, 2) {};
            \draw node at (-2.5, 2) {$v_2$};

            \draw node[dot=2mm] at (-2, 0) {};
            \draw node at (-2.5, 0) {$v_3$};
        \end{tikzpicture}
        \caption{Before extending}
        \label{fig:ExampleBefore}
    \end{subfigure}
    \hfill
    \begin{subfigure}[t]{0.4\linewidth}
        \begin{tikzpicture}
            \large
            \draw[draw=none,pattern=crosshatch] (2, 0) -- (0, 0) -- (0, 1) -- cycle;
            \draw[draw=none,pattern=horizontal lines] (0, 0) -- (-2, 0) -- (-2, 2) -- (0, 1) -- cycle;
            \draw (2, 0) -- (-2, 0);
            \draw (-2, 0) -- (-2, 2);
            \draw (-2, 2) -- (2, 0);
            \draw (0, 1) -- (0, 0);

            \draw[<->] (0, -0.5) -- (0, 2.5);

            \draw node[dot=2mm] at (2, 0) {};
            \draw node at (2.5, 0) {$v_1$};

            \draw node[dot=2mm] at (0, 1) {};
            \draw node at (0.5, 1) {$v_4$};

            \draw node[dot=2mm] at (-2, 2) {};
            \draw node at (-2.5, 2) {$v_2$};

            \draw node[dot=2mm] at (-2, 0) {};
            \draw node at (-2.5, 0) {$v_3$};

            \draw node[dot=2mm] at (0, 0) {};
            \draw node at (-0.3, -0.3) {$v_5$};
        \end{tikzpicture}
        \caption{After extending}
        \label{fig:ExampleAfter}
    \end{subfigure}
    \caption{Diagrams demonstrating the 2D $\extendname$ algorithm}
\end{figure}

\subsubsubsection{Example of the algorithm.}
Consider the polytope shown in~\pref{fig:ExampleBefore} with vertices $\{ v_1,
v_2, v_3 \}$, and suppose our activation function has two piecewise-linear
regions separated by the vertical line (1D hyperplane) $Nx + b = 0$ shown.
Because this hyperplane has some of the vertices of the polytope on one side
and some on the other, we will use it as the $N, b$ hyperplane on line 10. We
then find that $i$ on line 11 should be the last vertex on the first side of
the hyperplane, while $j$ should be the last vertex on the other side of the
hyperplane. We will assume things are oriented so that $i = v_1$ and $j = v_3$.

Then SplitPlane is called, which adds new vertices $p_i = v_4$ (shown
in~\pref{fig:ExampleAfter}) where the edge $v_1\to v_2$ intersects the
hyperplane, as well as $p_j = v_5$ where the edge $v_3 \to v_1$ intersects the
hyperplane.  Separating all of the vertices on the left of the hyperplane from
those on the right, we find that this has partitioned the original polytope
into two sub-polytopes, each on exactly one side of the hyperplane, as desired.
If there were more intersecting hyperplanes, we would then recurse on each of
the newly-generated polytopes to further subdivide them by the other
hyperplanes.

}{
    Please see~\meta{cite arXiv} for a worked example of the algorithm's
    execution.
}{}

\subsection{Representing Polytopes}
\label{sec:Algorithms-HvsV}

We close this section with a discussion of implementation concerns when
representing the convex polytopes that make up the partitioning of
$\hatr{f}{X}$. In standard computational geometry, bounded polytopes can be
represented in two equivalent forms:
\begin{enumerate}
    \item The \emph{half-space} or \emph{H-representation}, which encodes the
        polytope as an intersection of finitely-many half-spaces. (Each
        half-space being defined as a halfspace defined by an affine inequality
        $Ax \leq b$.)
    \item The \emph{vertex} or \emph{V-representation}, which encodes the
        polytope as a set of finitely many points; the polytope is then taken
        to be the convex hull of the points (i.e., smallest convex shape
        containing all of the points).
\end{enumerate}
Certain operations are more efficient when using one representation compared to
the other. For example, finding the intersection of two polytopes in an
H-representation can be done in linear time by concatenating their
representative half-spaces, but the same is not possible in V-representation.

There are two main operations on polytopes we need perform in our algorithms:
(i)~splitting a polytope with a hyperplane, and (ii)~applying an affine map to
all points in the polytope. In general, the first is more efficient in an
H-representation, while the latter is more efficient in a V-representation.
However, when restricted to two-dimensional polygons, the former is also
efficient in a V-representation, as demonstrated by~\pref{alg:split-plane},
helping to motivate our use of the V-representation in our algorithm.

Furthermore, the two polytope representations have different resiliency to
floating-point operations. In particular, H-representations for polytopes in
$\mathbb{R}^n$ are notoriously difficult to achieve high-precision with, because
the error introduced from using floating point numbers gets arbitrarily large as
one goes in a particular direction along any hyperplane face.  Ideally, we would
like the hyperplane to be most accurate in the region of the polytope itself,
which corresponds to choosing the magnitude of the norm vector correctly.
Unfortunately, to our knowledge, there is no efficient algorithm for computing
the ideal floating point H-representation of a polytope, although libraries such
as APRON~\citep{apronlib} are able to provide reasonable results for
low-dimensional spaces.  However, because neural networks utilize extremely
high-dimensional spaces (often hundreds or thousands of dimensions) and we wish
to iteratively apply our analysis, we find that errors from using floating-point
H-representations can quickly multiply and compound to become infeasible. By
contrast, floating-point inaccuracies in a V-representation are directly
interpretable as slightly misplacing the vertices of the polytope; no
``localization'' process is necessary to penalize inaccuracies close to the
polytope more than those far away from it.

Another difference is in the space complexity of the representation. In
general, H-representations can be more space-efficient for common shapes than
V-representations. However, when the polytope lies in a low-dimensional
subspace of a larger space, the V-representation is usually significantly more
efficient.

Thus, V-representations are a good choice for low-dimensionality polytopes
embedded in high-dimensional space, which is exactly what we need for analyzing
neural networks with two-dimensional \restrsname{}. This is why we designed our
algorithms to rely on $\Vert(X)$, so that they could be directly computed on a
V-representation.

\subsection{Extending to Higher-Dimensional Subsets of the Input Space}
The 2D algorithm described above can be seen as implementing the recursive case
of a more general, $n$-dimensional version of the algorithm that recurses on
each of the $(n-1)$-dimensional facets.
\onlyfor{arxiv}{
In 2D, we trace
the edges (1D faces) and use the 1D algorithm
from~\cite{DBLP:conf/nips/SotoudehT19} to subdivide them based on intersections
with the hyperplanes defining the function. More generally, for an arbitrary
$n$-dimensional polytope we can trace the $(n-1)$-dimensional \emph{facets} of
the polytope, recursively applying the $(n-1)$-dimensional variant of the
algorithm to split those facets according to the linear partitions of the
function.

We have experimented with such approaches, but found that the overhead of
keeping track of all $(n-k)$-dimensional faces (commonly known as the \emph{face
poset} or \emph{combinatorial structure}~\cite{polyfaq} of a polytope) was too
large in higher dimensions. The two-dimensional algorithm addresses this
concern by storing the combinatorial structure \emph{implicitly}, representing
2D polytopes by their vertices in counter-clockwise order, from which edges
correspond exactly to sequential vertices. To our knowledge, such a compact
representation allowing arbitrary $(n-k)$-dimensional faces to be read off is not
known for higher-dimensional polytopes. Nonetheless, we hope that extending our
algorithms to GPUs and other massively-parallel hardware may improve
performance to mitigate such overhead.
}{Please see~\meta{cite arXiv} for more details.}

\section{\syrenn{} tool} 
\label{sec:Tool}

This section provides more details about the design and implementation of our
tool, \syrenn{} (Symbolic Representations of Neural Networks), which computes
$\hatr{f}{X}$, where $f$ is a DNN using only piecewise-linear layers and $X$ is
a union of one- or two-dimensional polytopes. The tool is open-source;
it is available under the MIT license at
\url{https://github.com/95616ARG/SyReNN} and in the PyPI package
\texttt{pysyrenn}.

\subsubsubsection{Input and output format.}
\syrenn{} supports reading DNNs 
from two standard formats: ERAN (a textual format used by the ERAN
project~\cite{ERAN}) as well as ONNX (an industry-standard format supporting a
wide variety of different models)~\cite{onnx}.  Internally, the input DNN is described as
an instance of the \texttt{Network} class, which is itself a list of sequential
\texttt{Layer}s. A number of layer types are provided by \syrenn{}, including
\texttt{FullyConnectedLayer}, \texttt{ConvolutionalLayer}, and
\texttt{ReLULayer}. To support more complicated DNN architectures, we have
implemented a \texttt{ConcatLayer}, which represents a concatenation of the
output of two different layers.  The input region of interest, $X$, is defined
as a polytope described by a list of its vertices in counter-clockwise order.
The output of the tool is the symbolic representation~$\hatr{f}{X}$.

\subsubsubsection{Overall Architecture.}
We designed \syrenn{} in a client-server architecture using gRPC \cite{grpc} and
protocol buffers \cite{protobuf} as a standard method of communication between
the two. This architecture allows the bulk of the heavy computation to be done
in efficient C++ code, while allowing user-friendly interfaces in a variety of
languages. It also allows practitioners to run the server remotely on a more
powerful machine if necessary. The C++ server implementation uses the Intel TBB
library for parallelization. Our official front-end library is written in
Python, and available as a package on PyPI so installation is as simple as
\texttt{pip install pysyrenn}. The entire project can be built using the Bazel
build system, which manages dependencies using checksums.

\subsubsubsection{Server Architecture.}
The major algorithms are implemented as a gRPC server written in C++.
When a connection is first made, the server initializes the state with an empty
DNN $f(x) = x$. During the session, three operations are permitted: (i) append
a layer $g$ so that the current session's DNN is updated from $f_0$ to $f_1(x)
\coloneqq g(f_0(x))$, (ii) compute $\hatr{f}{X}$ for a one-dimensional $X$, or
(iii) compute $\hatr{f}{X}$ for a two-dimensional $X$.
We have separate methods for one- and two-dimensional $X$, because the
one-dimensional case has specific optimizations for controlling memory usage.
The \texttt{SegmentedLine} and \texttt{UPolytope} types are used to represent
one- and two-dimensional partitions of $X$, respectively.
When operation~(1) is performed, a new instance of the \texttt{LayerTransformer}
class is initialized with the relevant parameters and added to a running
\texttt{vector} of the current layers.  When operation~(2) is performed, a new
queue of \texttt{SegmentedLine}s is constructed, corresponding to $X$, and the
before-allocated \texttt{LayerTransformer}s are applied sequentially to compute
$\hatr{f}{X}$. In this case, extra control is provided to automatically gauge
memory usage and pause computation for portions of $X$ until more memory is made
available. Finally, when operation (3) is a performed, a new instance of
\texttt{UPolytope} is initialized with the vertices of $X$ and the
\texttt{LayerTransformer}s are again applied sequentially to compute
$\hatr{f}{X}$.

\subsubsubsection{Client Architecture.}
Our Python client exposes an interface for defining DNNs similar to the popular
Sequential-Network Keras API~\cite{keras}. Objects represent individual layers in the
network, and they can be combined sequentially into a \texttt{Network}
instance.  The key addition of our library is that this \texttt{Network}
exposes methods for computing $\hatr{f}{X}$ given a V-representation
description of $X$. To do this, it invokes the server and passes a
layer-by-layer description of $f$ followed by the polytope $X$, then parses the
response $\hatr{f}{X}$.

\subsubsubsection{Extending to support different layer types.}
Different layer types and activation functions are supported by sub-classing the
\texttt{LayerTransformer} class. Instances of \texttt{LayerTransformer}
expose a method for computing $\extendname{}(h, \cdot)$ for the
corresponding layer $h$. To simplify implementation, two sub-classes of
\texttt{LayerTransformer} are provided: one for entirely-linear layers (such as
fully-connected and convolutional layers), and one for piecewise-linear layers.
For fully-linear layers, all that needs to be provided is a method computing the
layer function itself. For piecewise-linear layers, two methods need to be
provided: (1)~computing the layer function itself, and (2)~one describing the
hyperplanes which separate the linear regions. The base class then directly
implements~\pref{alg:relu-2d-extend} for that layer. This architecture makes
supporting new layers a straight-forward process.

\subsubsubsection{Float Safety.}
Like Reluplex~\cite{reluplex:CAV2017}, \syrenn{} uses floating-point arithmetic
to compute $\hatr{f}{X}$ efficiently.  Unfortunately, this means that in some
cases its results will not be entirely precise when compared to a real-valued
or multiple-precision version of the algorithm.
\onlyfor{arxiv}{
If a perfectly precise solution is required, the server code can be modified to
use multiple-precision rationals instead of floats. Alternatively, a
confirmation pass can be run using multiple-precision numbers after the initial
float computation to confirm the accuracy of its results. The use of
over-approximations may also be explored for ensuring correctness with
floating-point evaluation, like in DeepPoly~\cite{Singh:POPL2019}.
Unfortunately, our algorithm does not directly lift to using such
approximations, since they may blow the originally-2D region into a
higher-dimensional (but very ``flat'') over-approximate polytope, preventing us
from applying the 2D algorithm for the next layer.
}{Approaches for addressing this are discussed in~\meta{cite arxiv}.}

\section{Applications of \syrenn{}}
\label{sec:Applications}

This section presents the use of \syrenn{} in three example case studies.

\subsection{Integrated Gradients}
\label{sec:IntegratedGradients}

A common problem in the field of \emph{explainable machine learning} is
understanding \emph{why} a DNN made the prediction it did. For example, given
an image classified by a DNN as a `cat,' why did the DNN decide it was a cat
instead of, say, a dog? Were there particular pixels which were particularly
important in deciding this?
Integrated Gradients (IG)~\cite{Sundararajan:ICML2017} is the state-of-the-art
method for computing such \emph{model attributions.}
\begin{definition}
    Given a DNN $f$, the \emph{integrated gradients along dimension $i$ for
    input $x$ and baseline $x'$} is defined to be:
    \begin{equation}
        \small
        IG_i(x)  \eqdef  (x_i - x_i') \times \int_{\alpha=0}^{1}\frac{ \partial f(x' + \alpha \times (x-x') )}{\partial x_i}d\alpha.
        \label{eq:IG}
    \end{equation}
\end{definition}
The computed value $IG_i(x)$ determines relatively how important the $i$th
input (e.g., pixel) was to the classification.

However, exactly computing this integral requires a symbolic, closed form for
the gradient of the network. Until~\cite{DBLP:conf/nips/SotoudehT19}, it was
not known how to compute such a closed-form and so IGs were always only
\emph{approximated} using a sampling-based approach. Unfortunately, because it
was unknown how to compute the true value, there was no way for practitioners
to determine how accurate their approximations were. This is particularly
concerning in fairness applications where an accurate attribution is
exceedingly important.

In~\cite{DBLP:conf/nips/SotoudehT19}, it was recognized that, when $X = \Hull(\{
    x, x' \})$, $\hatr{f}{X}$ can be used to \emph{exactly} compute $IG_i(x)$.
This is because within each partition of $\hatr{f}{X}$ the gradient of the
network is \emph{constant} because it behaves as a linear function, and hence
the integral can be written as the weighted sum of such finitely-many
gradients.\footnote{As noted in~\cite{DBLP:conf/nips/SotoudehT19}, this technically requires a
slight strengthening of the definition of $\hatr{f}{X}$ which is satisfied by
our algorithms as defined above.} Using our symbolic representation, the exact
IG can thus be computed as follows:
\begin{equation}
    \sum_{\Hull(\{ y_i, y'_i \}) \in \hatr{f}{\Hull(\{ x, x' \})}}
    (y'_i - y_i) \times
    \frac{\partial f(0.5\times (y_i + y'_i))}{\partial x_i}
    \label{eq:RIG}
\end{equation}
Where here $y_i, y'_i$ are the endpoints of the segment with $y_i$ closer to
$x$ and $y'_i$ closest to $x'$.

\subsubsubsection{Implementation.}
The helper class \texttt{IntegratedGradientsHelper} is provided by our Python
client library. It takes as input a DNN $f$ and a set of $(x, x')$
input-baseline pairs and then computes IG for each pair.
    
\subsubsubsection{Empirical Results.}
In~\cite{DBLP:conf/nips/SotoudehT19} \syrenn{} was used to show
conclusively that existing sampling-based methods were insufficient to
adequately approximate the true IG. This realization led to changes in the
official IG implementation to use the more-precise trapezoidal sampling method
we argued for.

\subsubsubsection{Timing Numbers.}
In those experiments, we used \syrenn{} to compute $\hatr{f}{X}$ for three
different DNNs $f$, namely the small, medium, and large convolutional models
from~\cite{ERAN}. For each DNN, we ran \syrenn{} on 100 one-dimensional lines.
The 100 calls to \syrenn{} completed in 20.8 seconds for the small model, 183.3
for the medium model, and 615.5 for the big model. Tests were performed on an
Intel Core i7-7820X CPU at 3.60GHz with 32GB of memory.

\subsection{Visualization of DNN Decision Boundaries}
\label{sec:Visualization}

\begin{figure}[t]
    \centering
    \begin{subfigure}[t]{.32\linewidth}
        \includegraphics[width=\linewidth]{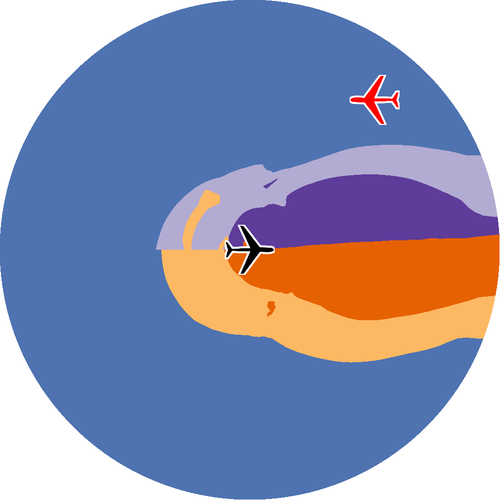}
        \caption{Decision boundaries computed using $\hatr{f}{X}$}
        \label{fig:netviz-fhat-all}
    \end{subfigure}
\hfill
    \begin{subfigure}[t]{.32\linewidth}
        \includegraphics[width=\linewidth]{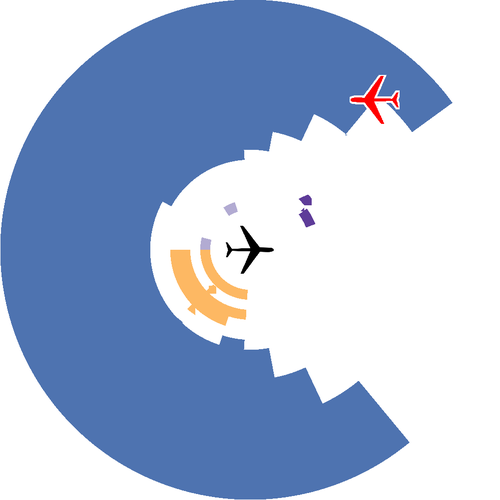}
        \caption{Decision boundaries computed using $\DPPre[k=25^2]$}
        \label{fig:netviz-dp25-all}
    \end{subfigure}
\hfill
    \begin{subfigure}[t]{.32\linewidth}
        \includegraphics[width=\linewidth]{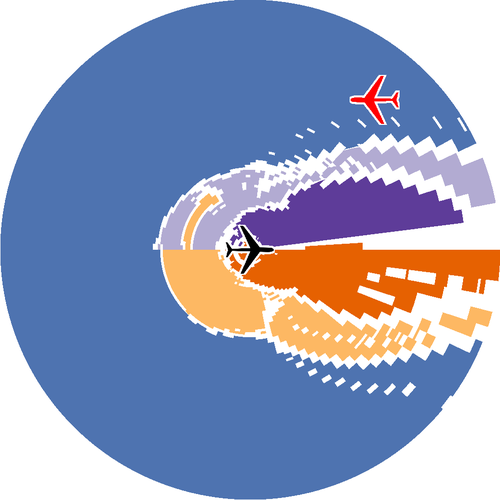}
        \caption{Decision boundaries computed using $\DPPre[k=100^2]$}
        \label{fig:netviz-dp100-all}
    \end{subfigure}
    \\
    {\smaller
    Legend: \textcolor{COCcolor}{\rule[.2\baselineskip]{1em}{2pt}} Clear-of-Conflict,
    \textcolor{WRcolor}{\rule[.2\baselineskip]{1em}{2pt}} Weak Right,
    \textcolor{SRcolor}{\rule[.2\baselineskip]{1em}{2pt}} Strong Right,
    \textcolor{SLcolor}{\rule[.2\baselineskip]{1em}{2pt}} Strong Left,
    \textcolor{WLcolor}{\rule[.2\baselineskip]{1em}{2pt}} Weak Left.}
    \caption{Visualization of decision boundaries for the ACAS Xu network.
    Using \syrenn{} (left) quickly produces the exact decision boundaries.
    Using abstract interpretation-based tools like DeepPoly (middle and right)
    are slower and produce only imprecise approximations of the decision
    boundaries.}
    \label{fig:eval-netviz-acas}
\end{figure}

\begin{table}[t]
    \centering
    \caption{Comparing the performance of DNN visualization using \syrenn{}
    versus DeepPoly for the ACAS Xu network~\citep{julian2018deep}.
    $\hatr{f}{X}$ size is the number of partitions in the symbolic
    representation. \syrenn{} time is the time taken to compute $\hatr{f}{X}$
    using \syrenn{}. $\DPPre[k]$ time is the time taken to compute DeepPoly for
    approximating decision boundaries with $k$ partitions. Each scenario
    represents a different two-dimensional slice of the input space; within
    each slice, the heading of the intruder relative to the ownship along with
    the speed of each involved plane is fixed.} {\small
    \begin{tabular}{@{}llllll@{}} \toprule
        & & & \multicolumn{3}{c}{$\DPPre$ time (secs)} \\
           \cmidrule{4-6}
        Scenario & $\hatr{f}{X}$ size & \syrenn{} time (secs) & k = $25^2$ & k = $55^2$ & k = $100^2$ \\ \midrule
        Head-On, Slow        & 33200 & 10.9 & 9.1 & 43.2 & 141.3 \\
        Head-On, Fast        & 30769 & 10.2 & 8.2 & 39.0 & 128.0 \\
        Perpendicular, Slow  & 37251 & 12.5 & 9.2 & 42.9 & 141.7 \\
        Perpendicular, Fast  & 33931 & 11.4 & 8.2 & 39.2 & 127.5 \\
        Opposite, Slow       & 36743 & 12.1 & 9.8 & 46.7 & 152.5 \\
        Opposite, Fast       & 38965 & 13.0 & 9.5 & 45.2 & 147.3 \\
        -Perpendicular, Slow & 36037 & 11.9 & 9.5 & 45.0 & 146.4 \\
        -Perpendicular, Fast & 33208 & 10.9 & 8.3 & 39.5 & 130.2 \\ \bottomrule
    \end{tabular}
    }
    \label{tab:eval-netviz-timing}
\end{table}

Whereas IG helps understand why a DNN made a particular prediction about a
single input point, another major task is \emph{visualizing} the decision
boundaries of a DNN on \emph{infinitely-many} input points.
\pref{fig:eval-netviz-acas} shows a visualization of an ACAS Xu
DNN~\cite{julian2018deep} which takes as input the position of an airplane and
an approaching attacker, then produces as output one of five advisories
instructing the plane, such as ``clear of conflict'' or to move ``weak left.''
Every point in the diagram represents the relative position of the approaching
plane, while the color indicates the advisory.

One approach to such visualizations is to simply sample finitely-many points
and extrapolate the behavior on the entire domain from those finitely-many
points. However, this approach is imprecise and risks missing vital information
because there is no way to know the correct sampling density to use to identify
all important features.

Another approach is to use a tool such as
DeepPoly~\cite{Singh:POPL2019} to over-approximate the output range of the DNN.
However, because DeepPoly is an over-approximation, there may be regions of the
input space for which it cannot state with confidence the decision made by the
network. In fact, the approximations used by DeepPoly are extremely coarse. A
na\"ive application of DeepPoly to this problem results in it being unable to make
claims about \emph{any} of the input space of interest. In order to utilize it,
we must \emph{partition} the space and run DeepPoly within each partition, which
significantly slows down the analysis.  Even when using $25^2$
partitions,~\pref{fig:netviz-dp25-all} shows that most of the interesting
region is still unclassifiable with DeepPoly (shown in white). Only when using
$100^2$ partitions is DeepPoly able to effectively approximate the decision
boundaries, although it is still quite imprecise.

By contrast, $\hatr{f}{X}$ can be used to \emph{exactly} determine the decision
boundaries on any 2D polytope subset of the input space, which can then be
plotted.  This is shown in~\pref{fig:netviz-fhat-all}.
Furthermore, as shown in~\pref{tab:eval-netviz-timing}, the approach using
$\hatr{f}{X}$ is \emph{significantly} faster than that using ERAN, even as we
get the \emph{precise} answer instead of an approximation.
Such visualizations can be particularly helpful in identifying issues to be
fixed using techniques such as those in~\pref{sec:Patching}.

\subsubsubsection{Implementation.}
The helper class \texttt{PlanesClassifier} is provided by our Python client
library.  It takes as input a DNN $f$ and an input region $X$, then computes
the decision boundaries of $f$ on $X$.

\subsubsubsection{Timing Numbers.}
Timing comparisons are given in~\pref{tab:eval-netviz-timing}. We see that
\syrenn{} is quite performant, and the exact \syrenn{} can be computed more
quickly than even a mediocre approximation from DeepPoly using $55^2$
partitions.
Tests were performed on a dedicated Amazon EC2 c5.metal instance, using
BenchExec~\cite{benchexec} to limit the number of CPU cores to 16 and RAM to
16GB.

\subsection{Patching of DNNs}
\label{sec:Patching}

\begin{figure}[t]
    \centering
    \begin{subfigure}[t]{.24\linewidth}
        \trimmedacas{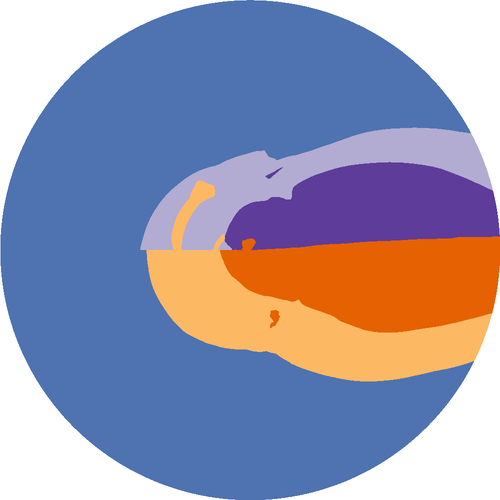}
        \caption{Before patching.}
        \label{fig:netpatch-pre}
    \end{subfigure}
    \hfill
    \begin{subfigure}[t]{.24\linewidth}
        \trimmedacas{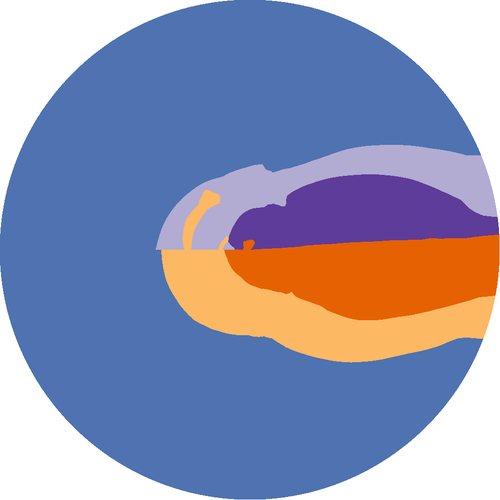}
        \caption{Patched pockets.}
        \label{fig:netpatch-pockets}
    \end{subfigure}
    \hfill
    \begin{subfigure}[t]{.24\linewidth}
        \trimmedacas{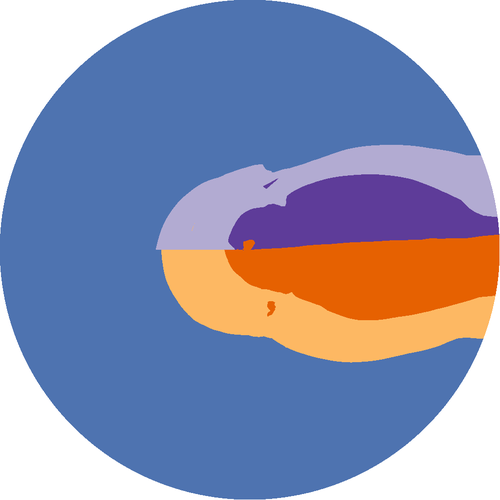}
        \caption{Patched bands.}
        \label{fig:netpatch-bands}
    \end{subfigure}
    \hfill
    \begin{subfigure}[t]{.24\linewidth}
        \trimmedacas{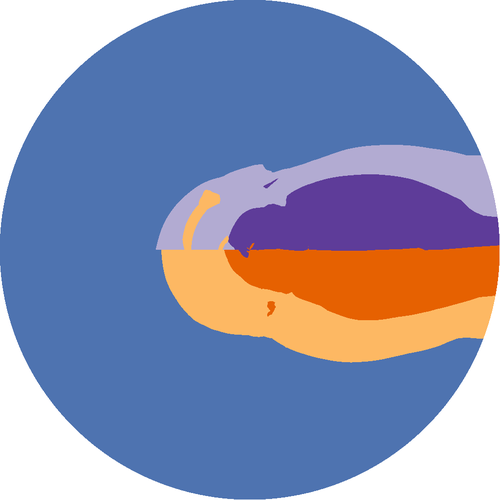}
        \caption{Patched symmetry.}
        \label{fig:netpatch-symmetry}
    \end{subfigure}
    \\
    {\smaller
    Legend: \textcolor{COCcolor}{\rule[.2\baselineskip]{1em}{2pt}} Clear-of-Conflict,
    \textcolor{WRcolor}{\rule[.2\baselineskip]{1em}{2pt}} Weak Right, 
    \textcolor{SRcolor}{\rule[.2\baselineskip]{1em}{2pt}} Strong Right, 
    \textcolor{SLcolor}{\rule[.2\baselineskip]{1em}{2pt}} Strong Left, 
    \textcolor{WLcolor}{\rule[.2\baselineskip]{1em}{2pt}} Weak Left.}
    \caption{Network patching.}
    \label{fig:netpatch}
\end{figure}

We have now seen how \syrenn{} can be used to visualize the behavior of a DNN.
This can be particularly useful for identifying buggy behavior. For example,
in~\pref{fig:netviz-fhat-all} we can see that the decision boundary between
``strong right'' and ``strong left'' is not symmetrical.

The final application we consider for \syrenn{} is \emph{patching DNNs} to
correct undesired behavior. Patching is described formally in~\cite{patching}.
Given an initial network $N$ and a \emph{specification} $\phi$ describing
desired constraints on the input/output, the goal of patching is to find a
small modification to the parameters of $N$ producing a new DNN $N'$ that
satisfies the constraints in $\phi$.

The key theory behind DNN patching we will use was developed
in~\cite{patching}. The key realization of that work is that, for a certain DNN
architecture, correcting the network behavior on an infinite, 2D region $X$ is
exactly equivalent to correcting its behavior on \emph{the finitely-many
vertices} $\Vert(P_i)$ for each of the finitely-many $P_i \in \hatr{f}{X}$.
Hence, \syrenn{} plays a key role in enabling efficient DNN patching.

For this case study, we patched the same aircraft collision-avoidance DNN
visualized in~\pref{sec:Visualization}. We patched the DNN three times to
correct three different buggy behaviors of the network: (i) remove ``Pockets''
of strong left/strong right in regions that are otherwise weak left/weak right;
(ii) remove the ``Bands'' of weak-left advisory behind and to the left of the
plane; and (iii) enforce ``Symmetry'' across the horizontal.  The DNNs before
and after patching with different specifications are shown
in~\pref{fig:netpatch}.

\subsubsubsection{Implementation}
The helper class \texttt{NetPatcher} is provided by our Python client
library. It takes as input a DNN $f$ and pairs of input region, output label
$X_i, Y_i$, then computes a new DNN $f'$ which maps all points in each
$X_i$ into $Y_i$.

\subsubsubsection{Timing Numbers.}
As in~\pref{sec:Visualization}, computing $\hatr{f}{X}$ for use in patching
took approximately 10 seconds.

\section{Related Work}
\label{sec:Related}

The related problem of exact reach set analysis for DNNs was investigated in
\cite{Xiang:arxiv2017}.  However, the authors use an algorithm that relies on
explicitly enumerating all exponentially-many ($2^n$) possible signs at each
\relu{} layer. By contrast, our algorithm adapts to the actual input polytopes,
efficiently restricting its consideration to activations that are actually
possible.

Hanin~and~Rolnick~\citet{regions2019icml} prove theoretical properties about
the cardinality of $\hatr{f}{X}$ for \relu{} networks, showing that
$\abs{\hatr{f}{X}}$ is expected to grow polynomially with the number of nodes
in the network for randomly-initialized networks.

Thrun~\citet{DBLP:conf/nips/Thrun94} and Bastani~et~al.\citet{DBLP:conf/nips/BastaniPS18} extract
symbolic rules meant to approximate DNNs, which can be thought of as an
approximation of the symbolic representation $\hatr{f}{X}$.

In particular, the ERAN~\cite{ERAN} tool and underlying
DeepPoly~\cite{Singh:POPL2019} domain were designed to verify the non-existence
of adversarial examples.
Breutel~et~al.~\citet{DBLP:conf/esann/BreutelMH03} presents an iterative refinement algorithm
that computes an overapproximation of the weakest precondition as a polytope
where the required output is also a polytope.

Scheibler~et~al.~\citet{DBLP:conf/mbmv/ScheiblerWWB15} verify the safety of a machine-learning
controller using the SMT-solver iSAT3, but support small unrolling depths and
basic safety properties. Zhu~et~al.~\citet{DBLP:conf/pldi/ZhuXMJ19} use a synthesis
procedure to generate a safe deterministic program that can enforce safety
conditions by monitoring the deployed DNN and preventing potentially unsafe
actions.  The presence of adversarial and fooling inputs for DNNs as well as
applications of DNNs in safety-critical systems has led to efforts to verify
and certify
DNNs~\cite{Bastani:NIPS2016,reluplex:CAV2017,Ehlers:ATVA2017,Huang:CAV2017,ai2:SP2018,Bunel:NIPS2018,Weng:ICML2018,Singh:POPL2019,Anderson:PLDI2019}.
\emph{Approximate reachability analysis} for neural networks safely
overapproximates the set of possible outputs
\citep{ai2:SP2018,Xiang:arxiv2017,Xiang:ACC2018,Weng:ICML2018,Dutta:NFM2018,ReluVal:Usenix2018}.

Prior work in the area of network patching focuses on enforcing constraints on
the network during training.  DiffAI~\citep{diffai2018} is an approach to train
neural networks that are certifiably robust to adversarial perturbations. DL2~\citep{fischer2019dl2} allows for training and querying neural networks with
logical constraints.

\section{Conclusion and Future Work}
\label{sec:Conclusion}

We presented \syrenn, a tool for understanding and analyzing DNNs. Given a
piecewise-linear network and a low-dimensional polytope subspace of the input
subspace, \syrenn{} computes a symbolic representation that decomposes the
behavior of the DNN into finitely-many linear functions. We showed how to
efficiently compute this representation, and presented the design of the
corresponding tool. We illustrated the utility of \syrenn{} on three
applications: computing exact IG, visualizing the behavior of DNNs, and patching
(repairing)~DNNs.

In contrast to prior work, \syrenn{} explores a unique point in the design
space of DNN analysis tools. In particular, instead of trading off precision of
the analysis for efficiency, \syrenn{} focuses on analyzing DNN behavior on
\emph{low-dimensional subspaces} of the domain, for which we can provide
\emph{both} efficiency and precision.

We plan on extending \syrenn{} to make use of GPUs and other massively-parallel
hardware to more quickly compute $\hatr{f}{X}$ for large $f$ or $X$. Techniques
to support input polytopes that are greater than two dimensional is also a ripe
area of future work. We may also be able to take advantage of the fact that
non-convex polytopes can be represented efficiently in 2D. Extending algorithms
for $\hatr{f}{X}$ to handle architectures such as Recurrent Neural Networks
(RNNs) will open up new application areas for \syrenn{}.

\paragraph{Acknowledgements.} We thank the anonymous reviewers for their
feedback and suggestions on this work. This material is based upon work
supported by a Facebook Probability and Programming award.

\bibliographystyle{splncs04}
\bibliography{main}

\vfill

{\small\medskip\noindent{\bf Open Access} This chapter is licensed under the terms of the Creative Commons\break Attribution 4.0 International License (\url{https://creativecommons.org/licenses/by/4.0/}), which permits use, sharing, adaptation, distribution and reproduction in any medium or format, as long as you give appropriate credit to the original author(s) and the source, provide a link to the Creative Commons license and indicate if changes were made.}

{\small \spaceskip .28em plus .1em minus .1em The images or other third party material in this chapter are included in the chapter's Creative Commons license, unless indicated otherwise in a credit line to the material.~If material is not included in the chapter's Creative Commons license and your intended\break use is not permitted by statutory regulation or exceeds the permitted use, you will need to obtain permission directly from the copyright holder.}

\medskip\noindent\includegraphics{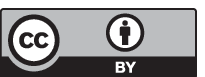}

\end{document}